\documentclass[a4paper,twoside]{article}

\usepackage{epsfig}
\usepackage{subfigure}
\usepackage{calc}
\usepackage{amssymb}
\usepackage{amstext}
\usepackage{amsmath}
\usepackage{amsthm}
\usepackage{multicol}
\usepackage{pslatex}
\usepackage{apalike}
\usepackage{SCITEPRESS}     

\begin{document}

\title{An adaptive thresholding approach for automatic optic disk segmentation }

\author{\authorname{F. Ghadiri\sup{1}, R. Bergevin\sup{1} and M. Shafiee \sup{2}}
\affiliation{\sup{1}Department of Electrical Engineering, Laval University, Quebec, Canada}
\affiliation{\sup{2}Ophthalmic Research Center,Khatam-al-Anbia Hospital, Mashhad University of Medical Sciences}
\email{\{farnoosh.ghadiri.1, robert.bergevin\}@ulaval.ca, hafieeM891@mums.ac.ir}
}

\keywords{Optic disk segmentation, Adaptive threshold, Vessel segmentation.}

\abstract{Optic disk segmentation is a prerequisite step in automatic retinal screening systems. In this paper, we propose an algorithm for optic disk segmentation based on a local adaptive thresholding method. Location of the optic disk is validated by intensity and average vessel width of retinal images. Then an adaptive thresholding is applied on the temporal and nasal part of the optic disc separately. Adaptive thresholding, makes our algorithm robust to illumination variations and  various image acquisition conditions. Moreover, experimental results on the DRIVE and KHATAM databases show promising results compared to the recent literature. In the DRIVE database, the optic disk in all images is correctly located and the mean overlap reached to 43.21\%. The optic disk is correctly detected in 98\% of the images with the mean overlap of 36.32\% in the KHATAM database.
}

\onecolumn \maketitle \normalsize \vfill

\section{\uppercase{Introduction}}
\label{sec:introduction}

\noindent Optic disk segmentation is one of the major steps in an automated retinal image analysis system. It is used as a prerequisite stage in the blood vessel tracking approaches to localize seed points and also optic disk segmentation improve the lesion diagnosis performance by qualifying lesion like exudate lesions. Moreover, optic disk segmentation helps ophthalmologists collect the information from optic disk such as color, size and shape in order to detect anomalies.

There are some characteristics which can be used to extract the optic disk. Sometimes, the optic disk area is brighter than its surrounding and it can be seen as a disk and in other images it appears as a hollow ring. Moreover, the optic disk part located on the temporal side is often more visible and brighter than the nasal part. 

Despite these features, optic disk segmentation is not an easy task because in some images the optic disk boundary is not well defined and it is obscured by the crossing blood vessels. In addition, the optic disk diameter can vary from 70 to 200 pixel in the DRIVE database. Therefore, proposing an algorithm which addresses these problems is essential.

In this paper, we present an algorithm for optic disk segmentation based on a thresholding approach. Initial candidates of the optic disk are chosen based on the average size of the retinal image. Then, the optic disk location is verified by choosing the region with high average vessel width. Finally, the optic disk is segmented by applying different thresholds on different sides of it.

This paper is organized as follows: In the first section, we review some state of art optic disk segmentation algorithms and give an overview of problems in this area. Section 2 introduces our algorithm in three parts: optic disk localization, vessel width computation, and optic disk segmentation. Section 3 presents our experimental results. Finally, further results are discussed in the conclusion section.

\section{\uppercase{Related Work}}

\noindent There have been several approaches for optic disk segmentation. The major efforts in this domain can be divided in two techniques; template-based methods (methods for obtaining optic disk boundary approximations) \cite{Aquino} and methods based on deformable models of snakes for extracting the optic disk boundary as exactly as possible.

\cite{Esmaeili} proposed an algorithm based on curvelet transform and deformable vibrational level set model. In this algorithm probable optic disk areas are obtained by applying curvelet transform and selecting the brightest area on the enhanced images. Then the region with the high value coefficients in the modified reconstructed image is choosen as the optic disk location. Finally, blood vessels are removed by morphological operation and the optic disk boundary extracted using level set deformable model.  This algorithm has two problems: first, applying curvelet transform and morphological operation to the entire image is time consuming. Also, using Starck et al. \cite{Starck} algorithm to modify the curvelet coefficient following by the constant threshold fails to detect reliable blood vessels, because of vessel-like patterns such as hemorrages and micro aneurysms. Therefore, making decision based on these results affect the final result. 

\cite{Osareh} proposed a method based on template matching. They used template matching to estimate the position of the optic disk. This position is also used to initialize points for the active contour. Then, they used morphological operation to remove vessels and improve the boundary around the optic disk. Finally, the optic disk boundary is extracted by applying Snake to the derived region. \cite{Malek} used an iterative thresholding algorithm to approximate the center of the optic disk. Then they applied the PCA (Principal Component Analysis) to the candidate regions obtained from the previous stage to find the final location of the optic disk. Finally, like  \cite{Osareh} a morphological method is used to eliminate blood vessels followed by applying an active contour to extract optic disk boundary. 

Active contour methods work on a gradient of the image and lock onto homogenous region enclosed by strong gradient information. Since the gradient along the optic disk boundary is not homogeneous, active contour methods should be modified or combined with other methods. \cite{Osareh} and \cite{Malek} used morphological operation to address this problem. However, using morphological operations can also blur and change the location of the optic disk boundary. Therefore, the result of optic disk boundary segmentation is not reliable.

\cite{Xu} introduced a modified active contour model based on estimating the optic disk boundary. All pixels are listed in descending order of gray-level; the top 13\% pixels are selected as the candidate region of the optic disk. Then, the final optic disk center and its boundary are estimated by distance transform based on edge map. Finally the Snake technique is applied based on the information obtained from the previous stage. The first estimation of the optic disk center tends to fall in the optic disk temporal part and this area is surrounded by blood vessel. Therefore, the obtained radius usually is an approximate radius of the optic disk temporal part. \cite{Hsiao} localized the optic disk center by illumination correction algorithm. Then, the radius of the optic disk is estimated by Canny edge detection followed by Hough transform. Finally, supervised gradient vector flow snake (SGVF snake)  is used to extract the optic disk boundary. Although the SGVF snake can be more powerful to reduce the influence of blood vessel occlusion, finding the imprecise estimation of optic disk boundary in the first stage is the problem that still remains in this method. The reason is that the divergence pattern of retinal blood vessels starting from optic disk can be modeled as a semi-circular shape \cite{Hoover}. Therefore, the circle drawn by circular Hough transform tends to fit with these vessels. \cite{Cheng} propose an algorithm based on superpixel classification. They first divide each image into superpixels. Then each pixel is classified as a disk or non-disk by computing center surround statistic from superpixels and unifying them with histograms.

In this paper we propose a new algorithm for optic disk segmentation based on adaptive thresholding. We use grey level information of the nasal and temporal parts of the optic disk to address the optic disk non-homogeneity problem. Moreover, we improve the accuracy of optic disk localization by adding the vessel width information as well as the overall intensity of the region. Applying a fast vessel segmentation algorithm on the small part of the optic disk followed by thresholding approach reduces the time complexity of our algorithm considerably.

\section{Proposed Method}
\noindent The main idea of our algorithm is based on the fact that the optic disk part located on the temporal side is more visible and brighter than the nasal part. Therefore, we consider the optic disk region as composed of two homogenous parts. The first step in our algorithm is to localize the optic disk using the intensity information and vessel width average. Then, the optic disk is divided in two regions based on the directional information of major vessels which are originating from the optic disk. Finally, the optic disk is segmented by applying an adaptive threshold on its different sides.

\subsection{Optic Disk Localization} 

\noindent In the retinal images, the average optic disk area varies from 13\% to 20\% of the whole image. We choose 13\% of the highest intensity pixels as the first estimation of the optic disk area. To reduce the number of candidate regions, we apply circular criterion to the obtained connected components from the previous stage to remove bright linear structure. This also reduces the time complexity. 

\begin{figure}
\center
{\subfigure[]{\includegraphics[width=.5\linewidth,keepaspectratio]{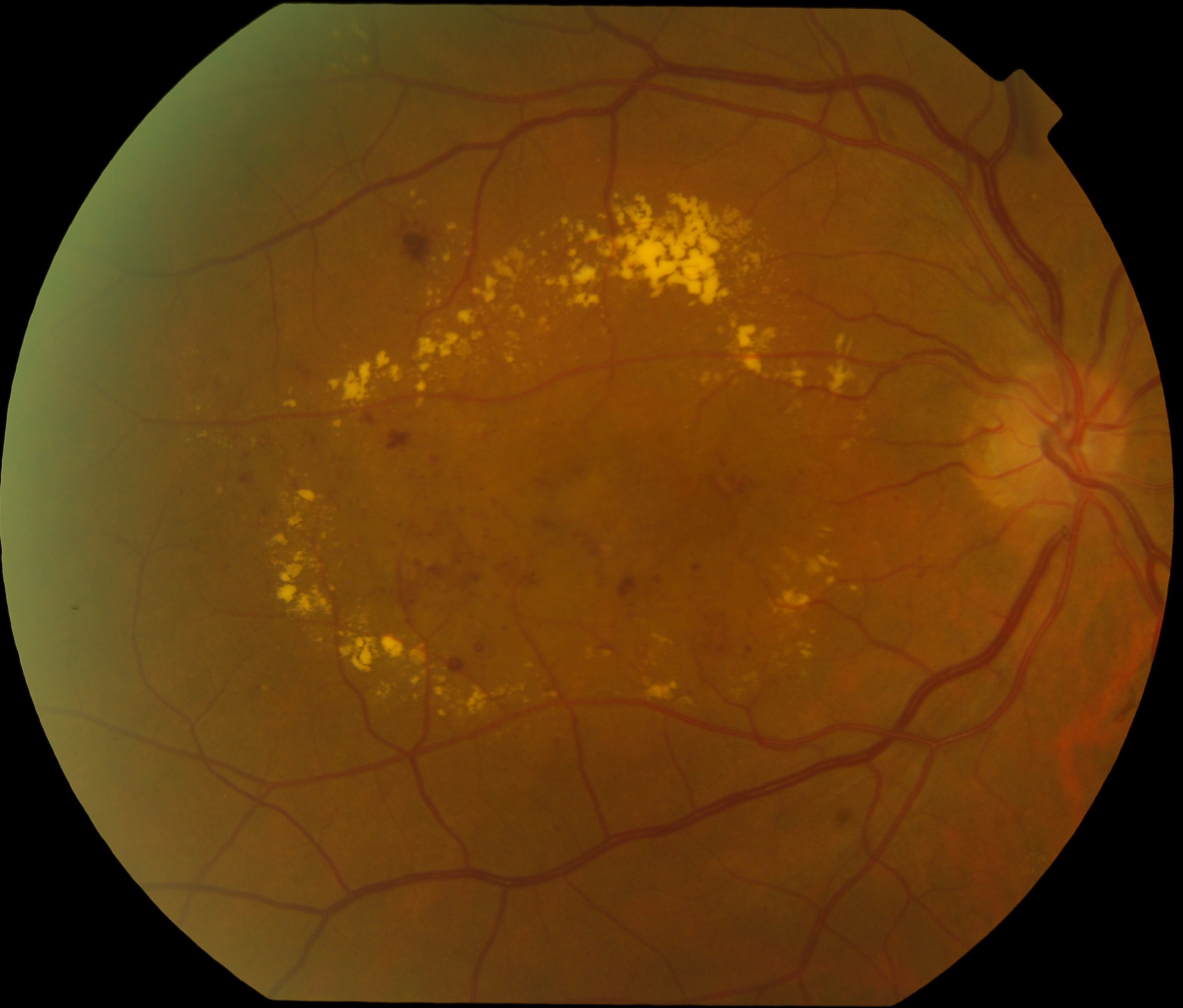}}}
{\subfigure[]{\includegraphics[width=.5\linewidth,keepaspectratio]{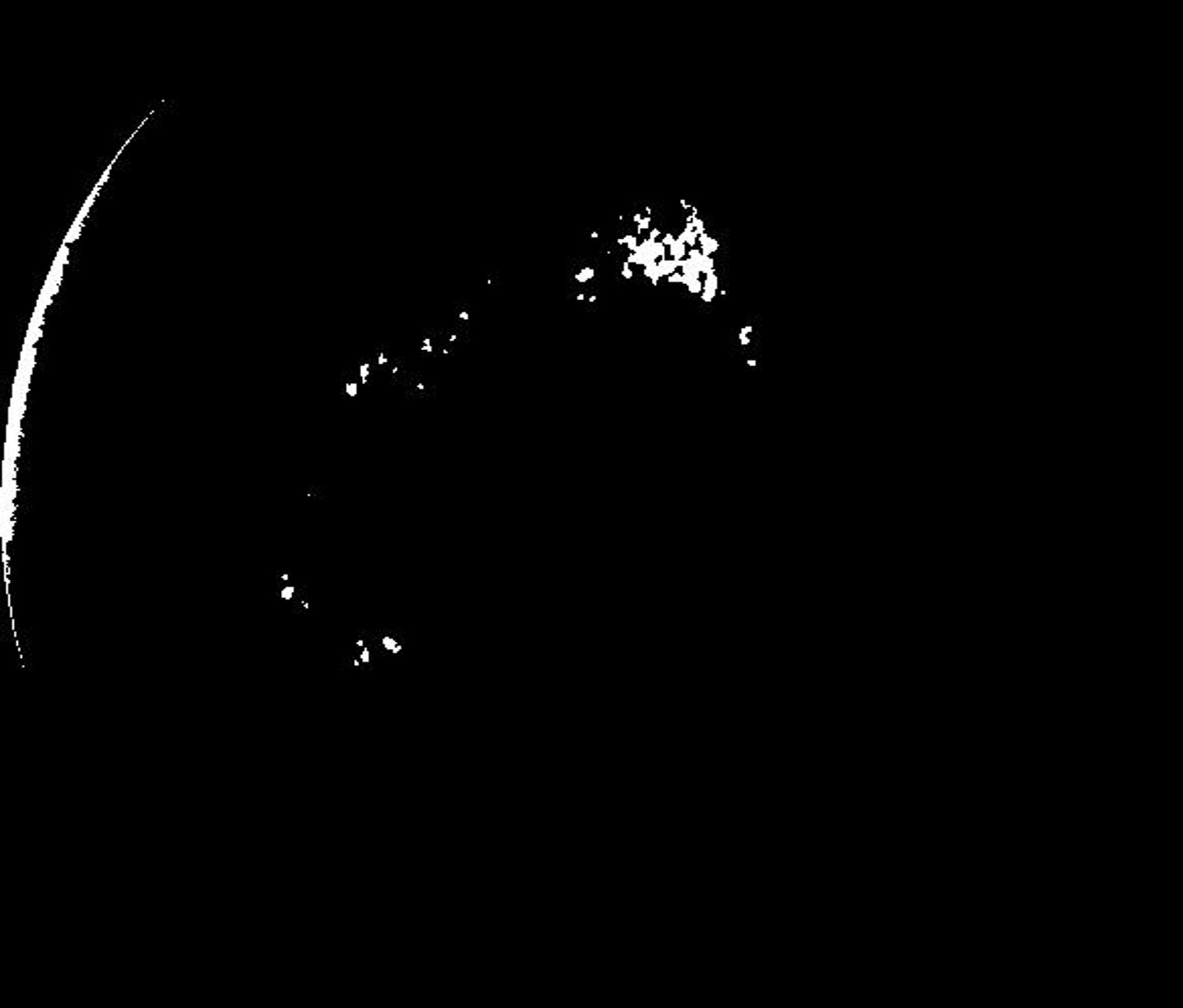}}}

{\subfigure[]{\includegraphics[width=.5\linewidth,keepaspectratio]{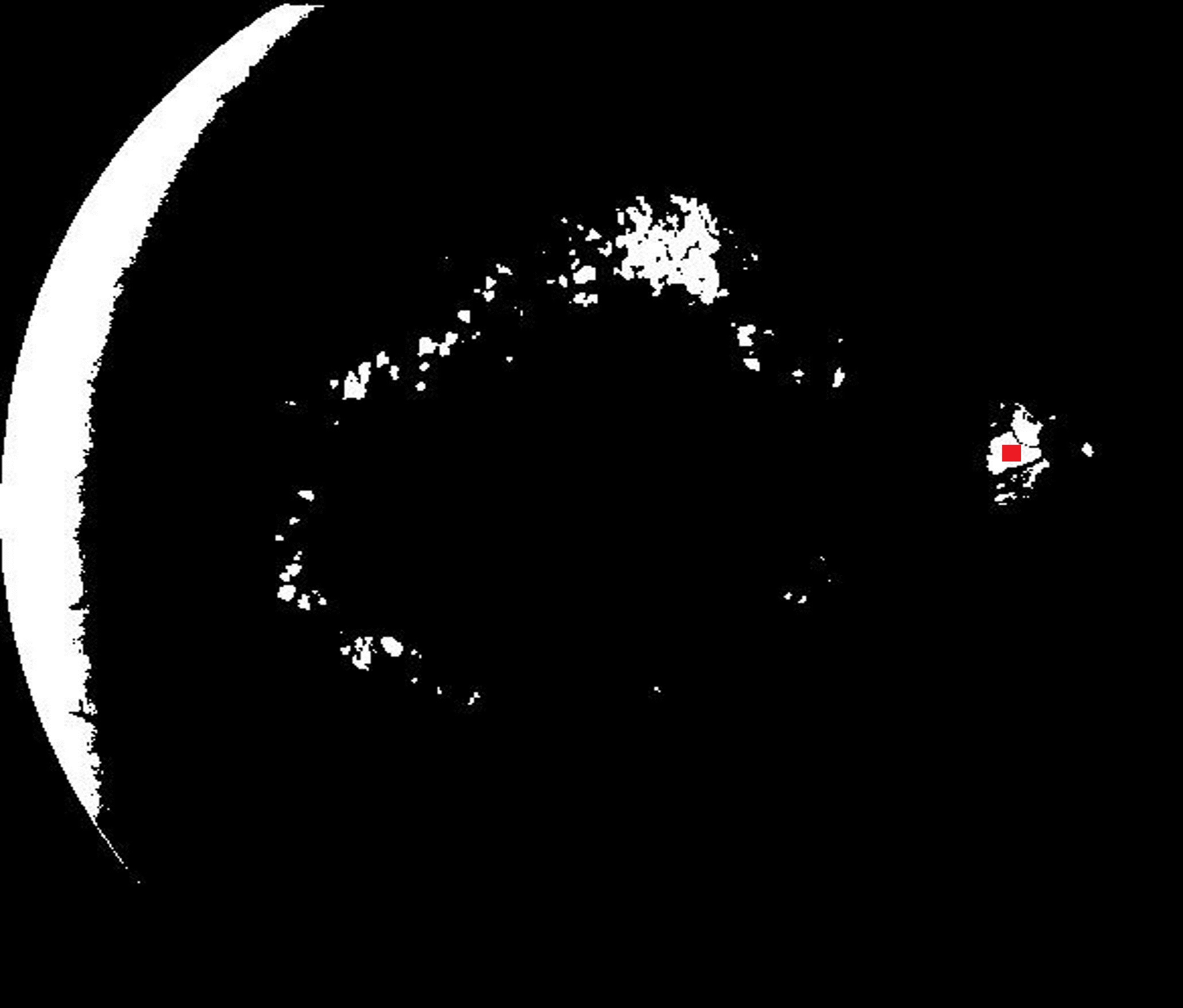}}}

\caption{(a) A retinal image form KHATAM database. (b) Initial candidates of optic disk after one iteration. (c) Initial candidates of optic disk after three iterations and the final optic disk location (red dot). 
}
\label{Tav2}%
\end{figure}

In the next step, the average blood vessel width is computed in the window at the center of each connected component, as further describe in section $3.2$. Then, the window with the highest average blood vessel width is selected as optic disk location. This value should be more than a fixed threshold, because in some cases, the optic disk area is not the brightest area of the retinal image. If no connected component has an average vessel width more than the defined threshold, we increase the threshold value of the first step and repeat the previous stages until circularity and average blood vessel width criterion are satisfied. Fig.\ref{Tav2} shows how the optic disk area appears after increasing the threshold value.

\subsection{Vessel width computation}
\noindent The optic disk is considered as a bright area where blood vessels converge. Average blood vessel width in the window around the optic disk area is larger than in other parts of the retinal image. Therefore, we use this feature to validate the optic disk location among the obtained candidates. To analyse blood vessel width in the candidate regions, we use \cite{Ghadiri} vessel segmentation algorithm.

In this algorithm, each region is divided into overlapping windows both horizontally and vertically. Then the direction of the linear structure in each window is computed by Radon transform. Afterwards the Sobel operator is applied along the obtained direction. The gradient information obtained from Sobel operator combined with the Fuzzy inference engine is used to decide whether the obtained gradient is related to the blood vessel edge or not. When of vessel edges are validated, blood vessel width is computed using the average distance between pairs of edges. Finally, validated vessels are reconstructed using morphological operations.

\subsection{Optic disk Segmentation}

\noindent To segment the optic disk we need to find the window that covers the whole optic disk area (circumscribed rectangle). To accomplish this task, we propose an algorithm composed of the four following steps.

\begin{itemize}
\item Finding the optic disk center.
\item  Detecting points on the optic disc boundary to estimate the circumscribed window around the optic disk.
\item Dividing the optic disk in two regions based on the its vessels structure.
\item Applying two different thresholds on these regions.
\end{itemize}

In the first step, we need to define a window with the center of the optic disk. To accomplish this task, the center of the obtained region from the optic disk localization step is modified. The obtained region from the optic disk localization step is located at the temporal part of the optic disk. The reason is that this part is brighter than the other parts of the optic disk (superior, inferior and nasal parts. Fig 2), so the center of this area can approximate the center of optic disk height. But, this point can be the center of optic disk width if only the brightest part of the temporal side is close to the optic disk vessels. Therefore, by the information obtained from the vessel segmentation step, we adjust this point so that it gets close to the vessel. This point is used as the first estimation of optic disk center.

In the second step, we assume a line which passes through the center of the optic disk. We apply Sobel operator in the perpendicular direction of this line. Center of the Sobel mask is chosen from the points close to the line’s end-points. 
From each side of the optic disk center the points whose gradient values are above a fixed threshold are selected. Among these point those which are labeled as vessel from the vessel segmentation step are removd. Finally, there should remain one point on each side of the center point with gradient signs opposite to each other. If the optic disk area is inscribed in the pre-defined window, we will have two points of optic disk boundary on each side of optic disk center. If the points are not found, the size of the window is slightly increased and the previous steps are repeated. Then, we repeat the first step to find the new points. When the points are found, the center of the optic disk is re-calculated based on the middle of these points. New center point is considered as the center of the circumscribing rectangle of the optic disk. The width of the rectangle is equal to the distance between the two points and its height is computed based on min to max disk diameter ratio in normal eyes \cite{Jost}.

In the third step, we divide optic disk area in two regions based on the major vertical vessels in the circumscribing rectangle. Our goal is to divide optic disk region into temporal and nasal parts. We use information obtained from the vessel width computing step described in $3.2$ to detect major vertical vessels. In the last step Then we divide the optic disk in two regions and apply an Otsu’s thresholding method threshold on each region. 

\begin{figure}[!h]
		\center
{\includegraphics[width=38mm]{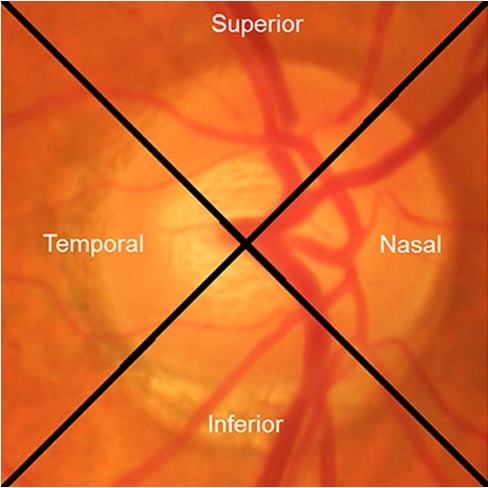}}
\caption{Annotations describe the four different zones of the optic disk \cite{Lu}.}
\end{figure}

\section{Experimental Results}
\noindent Our algorithm is basically defined by three main steps: optic disk localization, blood vessel segmentation and optic disk segmentation. In the first step, optic disk candidates are detected by a thresholding algorithm. Average vessel width is computed in a window size of 70*70 at the center of each candidate location. The window size is chosen based on the normal size of the optic disk. The candidate with the highest average vessel width is an input for the optic disk segmentation step. In the final step, optic disk is divided in two regions and Otsu’s thresholding method is applied on each region.

We applied our algorithm on 40 images of DRIVE dataset\footnote{http://www.isi.uu.nl/Research/Databases/DRIVE/}and 50 images of Khatam database\footnote{Images of this database are gathered by Khatam-Al-Anbia eye
hospital, Mashhad, Iran}. In both datasets the optic disk boundary is hand labeled by an ophthalmologist as ground truth. Fig. 3 and Fig. 4 show our results on the DRIVE and Khatam databases respectively. The results show that our algorithm is robust to changing illumination and also to abnormal retinal images. The result on the second image of Khatam dataset (Fig. 3) shows that our algorithm can distinguish between parapapillary atrophy boundary and optic disk boundary.

For the quantitative evaluation of our approach, we use sensitivity, specificity and overlap as follows:

\begin{equation}
\begin{gathered}
  sensitivity=\frac{TP}{(TP+FN)},\\
  specificity=\frac{TN}{(TN+FP)},\\
  overlap=\frac{area(A\cap B)}{area (B\cup B)}.
\end{gathered}
\end{equation}

TP, FN, TN and FP indicate true positive, false negative, true negative and false positive pixels respectively. Moreover, A and B are the ground truth region and optic disk region extracted from our algorithm, respectively.
Table 1 shows the results of these qualitative evaluation on the DRIVE database and Khatam database. We compare our algorithm to our implementation of \cite{Malek} and\cite{Xu}, we only consider those set of images in both datasets which their optic disk is detected in these three approaches. Table 2 shows the success rate of our algorithm in optic disk localization campare to the \cite{Malek} and\cite{Xu}. The results show that our method outperform the result presented in \cite{Malek} and \cite{Xu}.
\begin{figure}[!h]
	\center
{\includegraphics[width=30mm]{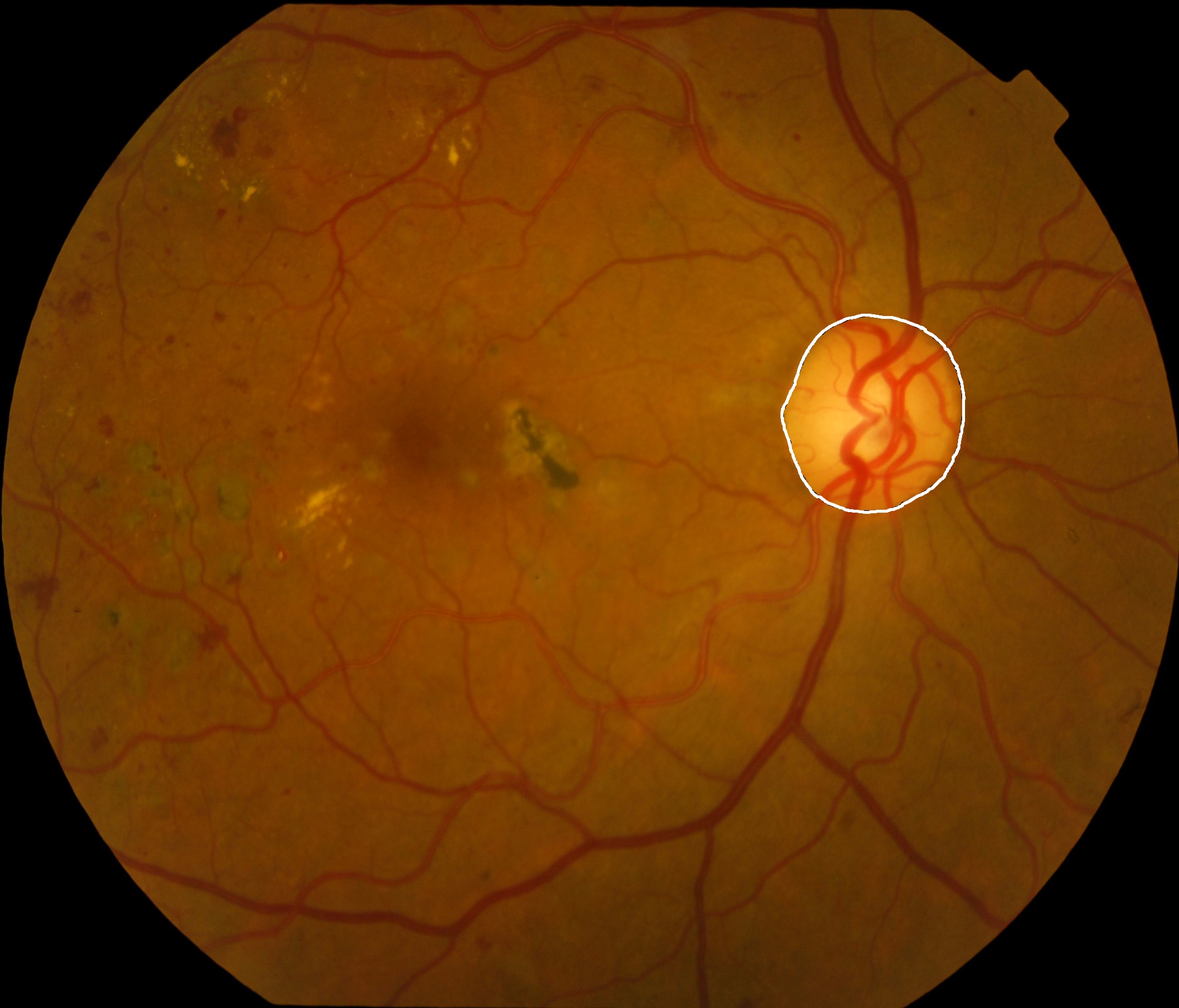}}
{\includegraphics[width=30mm]{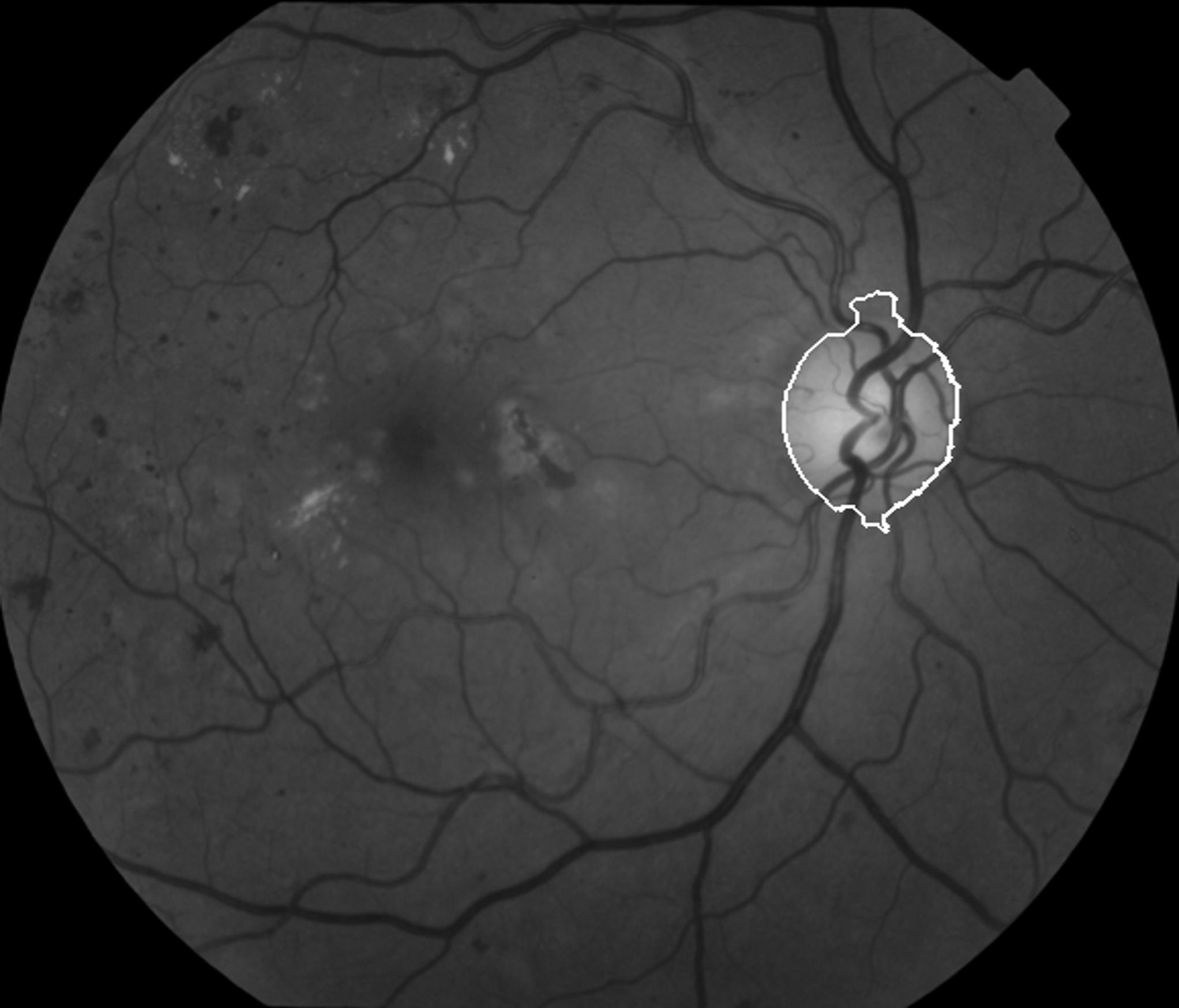}}
{\includegraphics[width=30mm]{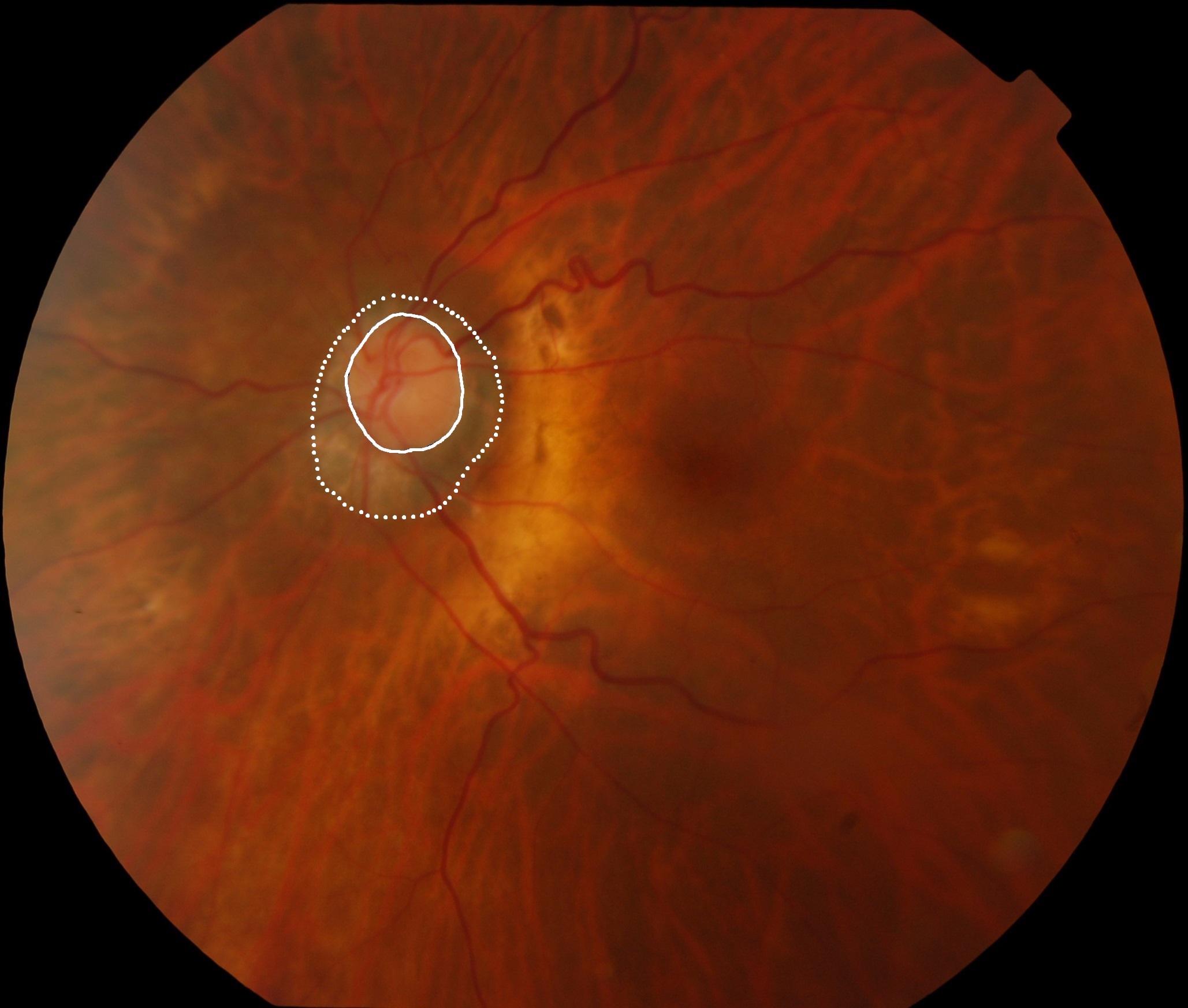}}
{\includegraphics[width=30mm]{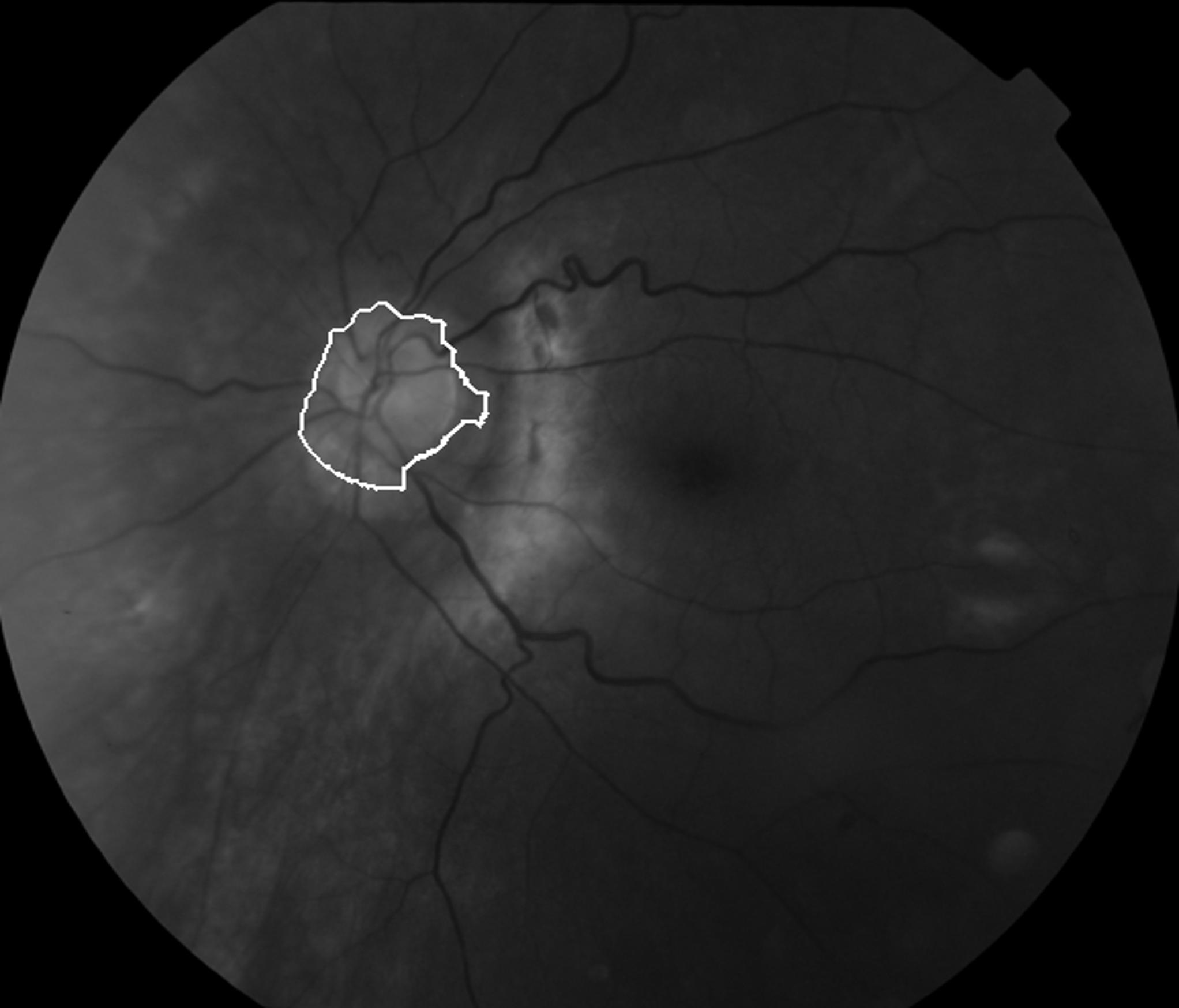}}
{\includegraphics[width=30mm]{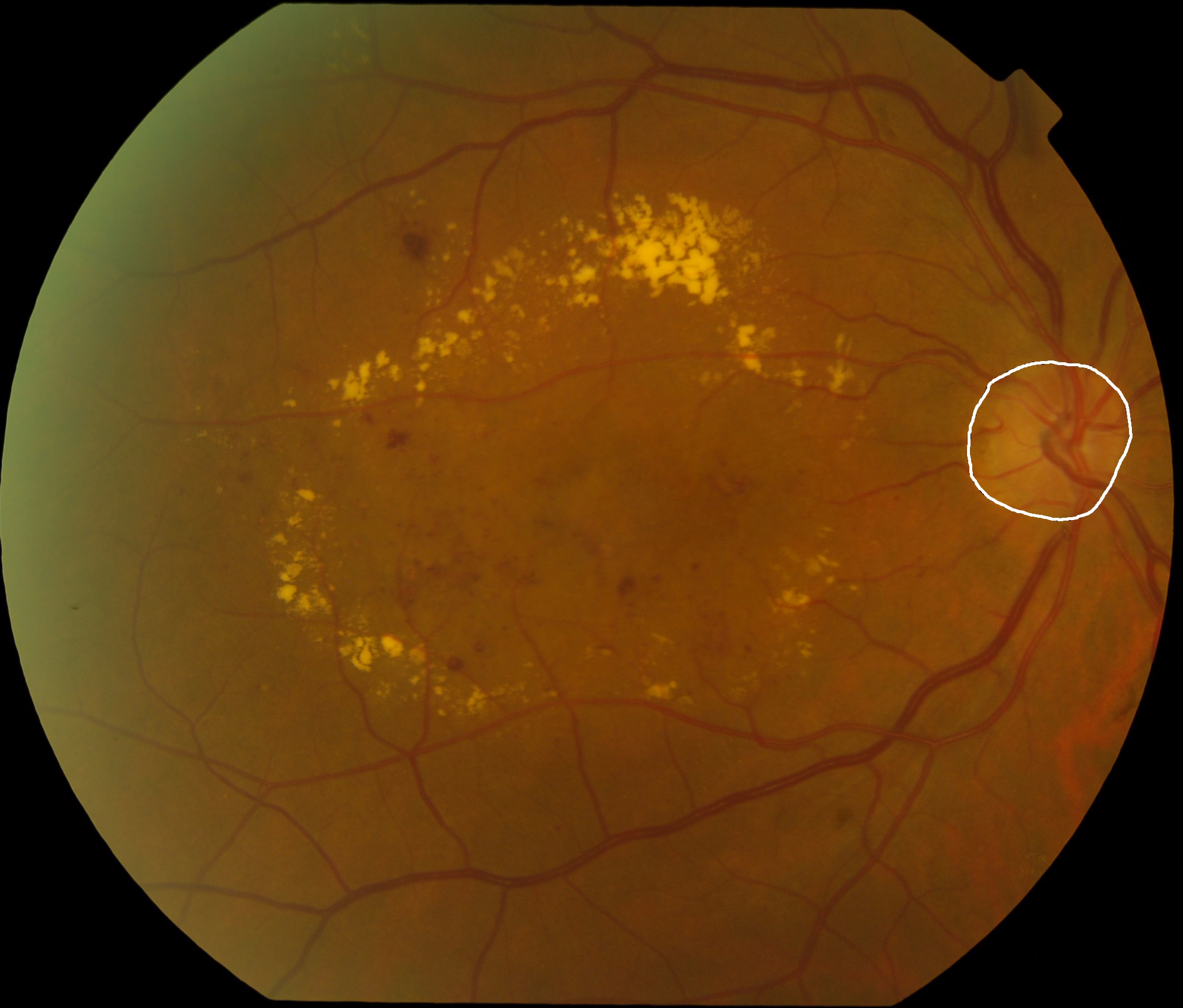}} 
{\includegraphics[width=30mm]{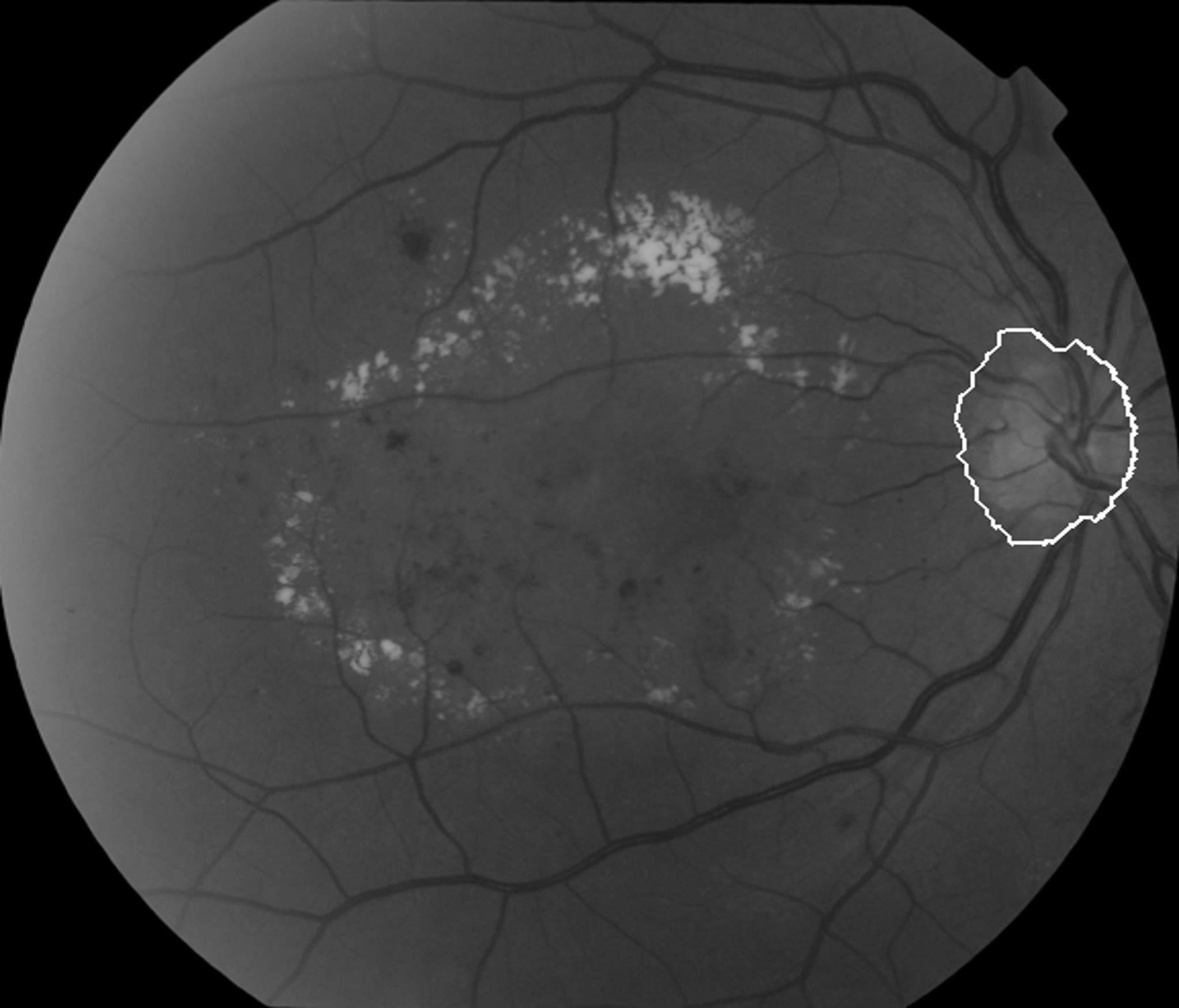}}

\caption{Optic disk segmentation results on three images of the Khatam database. \textbf{Left column:} Optic disk boundary detected by the proposed algorithm; parapapillary atrophy region is also shown in the second image of the left column with white dots. \textbf{Right column:} Optic disk boundary hand labeled by an ophtalmologist.}
\end{figure}
\begin{figure}
	\center
{\includegraphics[width=30mm]{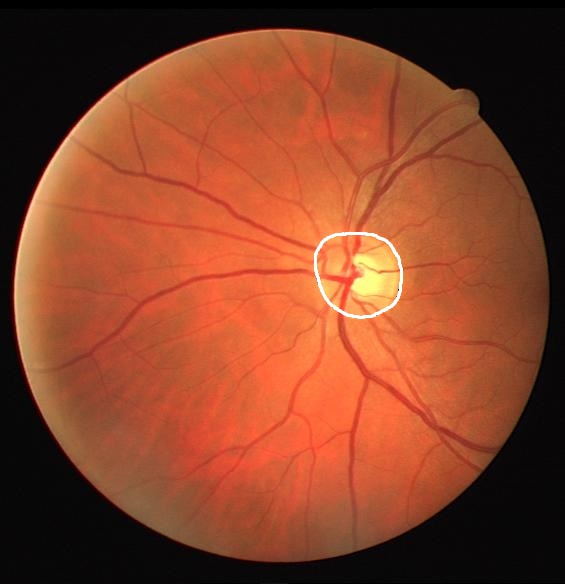}} 
{\includegraphics[width=30.02mm]{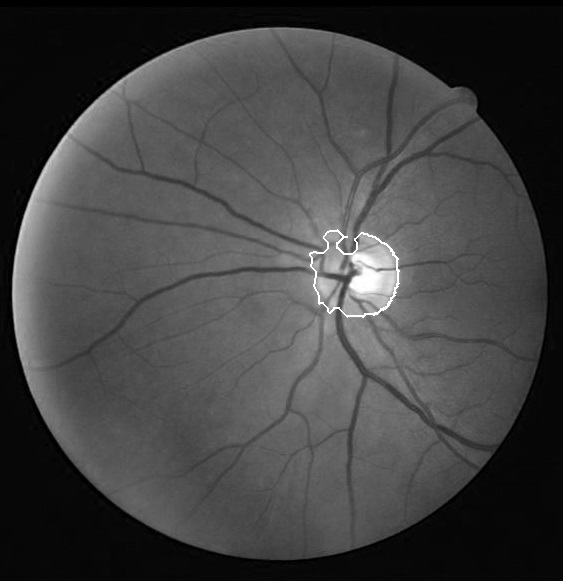}}
{\includegraphics[width=30mm]{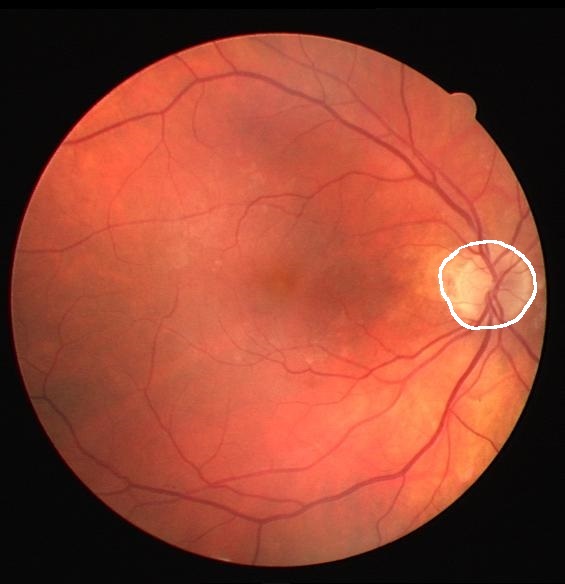}}
{\includegraphics[width=30mm]{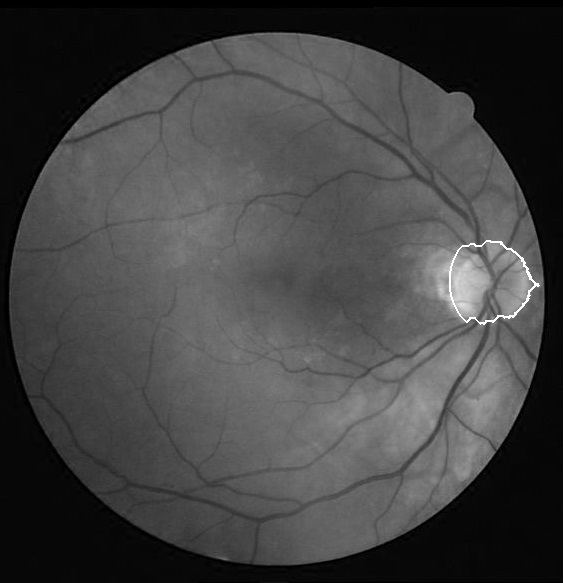}}
{\includegraphics[width=30mm]{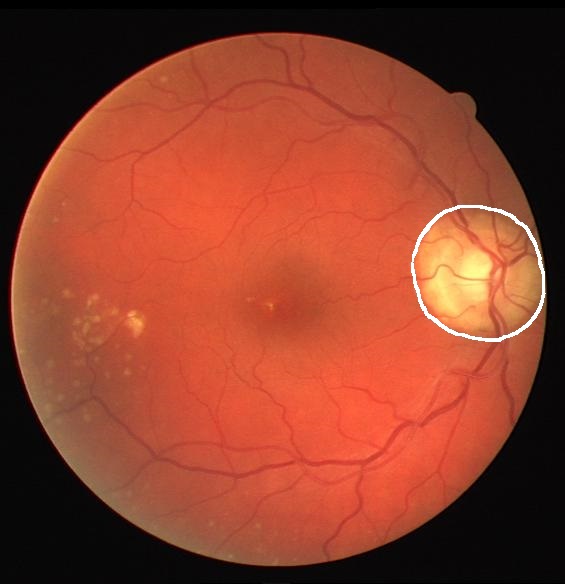}} 
{\includegraphics[width=30mm]{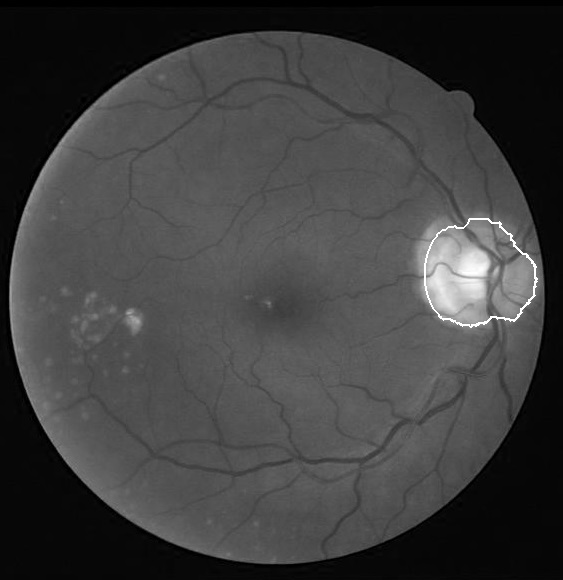}}

\caption{Optic disk boundary detection results on the three images of the DRIVE database. \textbf{Left column:} Optic disk boundary detected by the proposed algorithm. \textbf{Right column:} Optic disk boundary hand labeled by an ophtalmologist.}
\end{figure}

\setlength{\tabcolsep}{3pt}
\begin{table*}[t]
\centering
\caption{Comparative average sensitivity, specificity, overlap, and computation time on DRIVE and KHATAM dataset.}
\label{table:headings}
\begin{tabular}{llllllll}
\hline\noalign{\smallskip}
Methods &  Average  & Average  & Average  & Average time \\
  & sensitivity (\%) & specificity (\%) & overlap (\%) & per image (s) \\
\noalign{\smallskip}
\hline
\noalign{\smallskip}
DRIVE database  &  &   &  &   &\\
\cite{Malek} & 69.99 & 98.88 &33.66&  22.78&\\
\cite{Xu} & 73.68 & 99.20 &37.42 & 16.04 &\\
Our proposed method &80.46 & 99.72 & 43.21 &8.38 &\\
\\
Khatam database   &  &  &  &  &\\
\cite{Malek}  &73.45 &99.86 &29.45 &23.65&\\
\cite{Xu} & 77.12 &99.80& 30.18 &20.34&\\
Our proposed method & 82.46 & 99.81 & 36.32 & 16.58 &\\

\hline
\end{tabular}
\end{table*}

\begin{table*}[t]
\centering
\caption{Success rates of our algorithm versus \cite{Malek} and \cite{Xu} on optic disk localization.}
\label{table:headings}
\begin{tabular}{ccccc}
\hline\noalign{\smallskip}
Database &   \cite{Malek}  & \cite{Xu}  & Our proposed method \\
\noalign{\smallskip}
\hline
\noalign{\smallskip}
DRIVE  & 100 & 100 & 100 & \\
Khatam  & 94 & 92 & 98 & \\
\hline
\end{tabular}
\end{table*}

\section{Conclusion}
\noindent In this paper, we propose an algorithm for optic disk segmentation based on the intensity information of the optic disk and major vessels information. Experimental results show that our algorithm is superior to state of the art optic disk segmentation algorithms in terms of computation time and accuracy.

Using an iterative thresholding algorithm to find the initial candidates of the optic disk and adaptive thresholding for optic disk segmentation make our algorithm more robust to changing illumination. Moreover, vessel width information is used to reduce false positive detection of optic disk. Furthermore, computation time is reduced by using a simple thresholding algorithm and applying costly parts of the algorithm on a small area rather than the whole image.

\bibliographystyle{apalike}
{\small
\bibliography{example}}

\end{document}